# Automatic Classification of Bright Retinal Lesions via Deep Network Features

**Ibrahim Sadek**[a,b,*], **Mohamed Elawady**[c,e], **Abd El Rahman Shabayek**[d,e]
[a]Image and Pervasive Access Lab, CNRS UMI 2955, Singapore
[b]Télécom ParisTech, Institut Mines Télécom, France
[c]Universite Jean Monnet, CNRS, UMR 5516, Laboratoire Hubert Curien, F-42000, Saint-Etienne, France
[d]Interdisciplinary Centre for Security, Reliability, and Trust, University of Luxembourg, Luxembourg
[e]Department of Computer Science, Faculty of Computers and Informatics, Suez Canal University, Ismailia, Egypt

**Abstract.** The diabetic retinopathy is timely diagonalized through color eye fundus images by experienced ophthalmologists, in order to recognize potential retinal features and identify early-blindness cases. In this paper, it is proposed to extract deep features from the last fully-connected layer of, four different, pre-trained convolutional neural networks. These features are then fed into a non-linear classifier to discriminate three-class diabetic cases, i.e., normal, exudates, and drusen. Averaged across 1113 color retinal images collected from six publicly available annotated datasets, the deep features approach perform better than the classical bag-of-words approach. The proposed approaches have an average accuracy between 91.23% and 92.00% with more than 13% improvement over the traditional state of art methods.

**Keywords:** Diabetic Retinopathy, Exudates, Drusen, Bag of Words, Convolutional Neural Networks, Support Vector Machine. *Ibrahim Sadek, ibrahim.sadek@ipal.cnrs.fr

## 1 Introduction

Diabetic retinopathy (DR) and age-related macular degeneration (ARMD) are among the leading causes of visual impairment worldwide.[1] Hence, their detection is essential for diabetic retinopathy screening systems. Automated screening can significantly improve the disease identification and suggestion of early medical action. In this paper, we firstly show the application of deep transfer learning to discriminate a retinal image over non-binary classes problem (normal and two diseases). We used pre-defined deep features from popular CNN architectures that were employed in the tasks of ImageNet classification and recognition. Our work shows significant improvement over hand-crafted features. It needs a single feed-forward network pass followed by a non-linear classifier, in order to classify the retinal diseases based on the textural information of the vessels and the optic disc. The experimental results shows clear performance enhancement over the classical state of art bag-of-words method after being tested on six publicly available annotated datasets. To the best of our knowledge, this is the first work to 1) automatically extract retinal features from four different pre-trained deeply learned architectures (ResNet, GoogLeNet, VGG-VD and VGG), 2) classify three-classes of retinal images (normal, exudates, and drusen), 3) train and test retinal images with cross-validation with largely variant six different datasets.

The rest of the paper is organized as follows: Section 2 briefly explains the different cases of diabetic retinopathy and Section 3 gives an overview of related work. The proposed method is explained in Section 4 and the exprimental validation and results disccussion is covered in Section 5 and the work is finally concluded in Section 6.

## 2 Diabetic Retinopathy

Adults aged between 20 and 74 years are frequently diseased by Diabetic Retinopathy (DR), which is mainly identified by red lesions, i.e., microaneurysms and bright lesions, i.e., exudates that exist in the retina outer layer. Exudates have sharp margins and variable white/yellowish white shapes. People over 50 years are usually diseased by age-related macular degeneration (ARMD). It is designated by drusen, which appear as very small white/yellow deposits in the Bruch's membrane. The identification of lesions must be accurate to be useful for clinical application. Our focus will be on retinal lesions which appear as yellow-white spots such as exudates and drusen. Fig. 1 shows examples of retinal fundus images with bright lesions. The clinical lesions of the retina, i.e., cotton wool spots, hard exudates, and drusen are summarized as follows:[2]

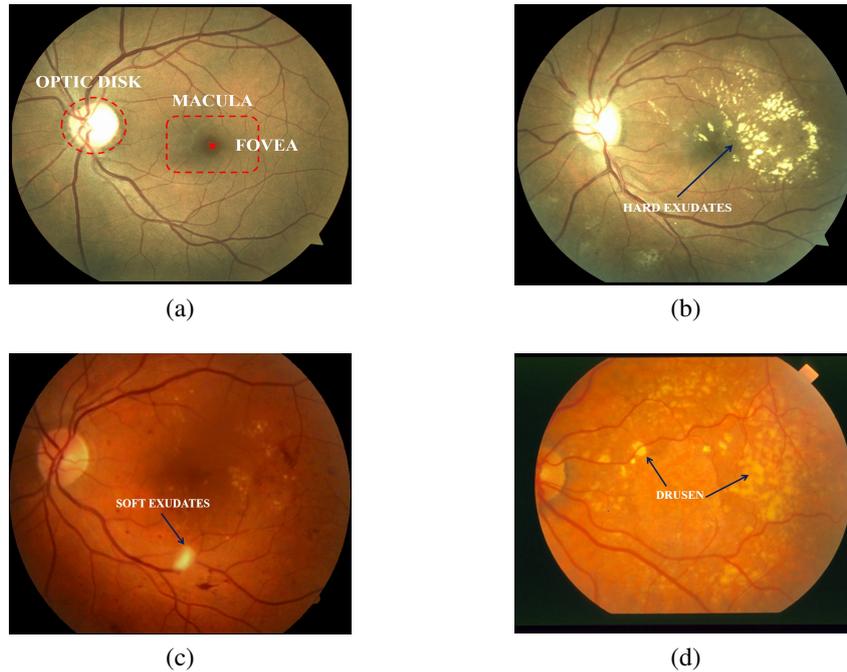

Fig 1: Typical fundus images; (a) normal, (b) hard Exudates, (c) soft exudates, (d) drusen.

*2.1 Soft Exudates or Cotton Wool Spots*

Soft Exudates or Cotton Wool Spots (CWS) appear as white, feathery spots with fuzzy borders. CWS physically correspond to small retinal closures (infarcts) and swellings of the retinal nerve fiber layer because of microvascular diseases. They are located in the superficial (inner) retina, so they may obscure nearby vessels. Fig. 1(c) shows an example of such lesions.

*2.2 Hard Exudates*

Hard Exudates (HE) are lipoprotein and other kinds of protein originating from leaking microaneurysms. HE appears as small white or yellowish white deposits with sharp edges, irregular shape, and variable size as presented in Fig. 1(b). The physical locations of these lesions are deeper in the retina than cotton wool spots.

*2.3 Drusen*

Drusen are variable sized yellowish lipoproteinaceous deposits that form between the retinal pigmented epithelium (RPE), and Bruch's membrane. Usually, drusen alone do not contribute to vision loss. However, an increase in the size or number of such lesions are the earliest signs of ARMD. Fig. 1(d) shows an example.

## 3 Previous Work

*3.1 Bag of Visual Words*

Recently, the bag of visual words approach (BoVW) has been widely used by many researchers to discriminates color retinal fundus images with or without diabetic retinopathy signs. In[3–5] bright (exudates, cotton wool spots, and drusen) and red (microaneurysms and hemorrhages) lesions are identified by extracting local feature descriptors from manually annotated regions of interest (ROIs). The authors provided a direct method in[6] to circumvent the manual selection of ROIs. Discrimination of bright lesions (drusen and exudates) is introduced in[7–9] by extracting local feature descriptors and color histogram features (RGB, HSV, and YCbCr color spaces) from local image patches. In[9] the patch location with respect to the image origin is required for the classification process. A pixel-based approach is discussed in[10] to identify images with no lesions and images with drusen or exudates by extracting shape, context, intensity, color, and statistical features from the segmented lesions.

Apart from the extracted features, the difference between the above-mentioned methods arises from feature encoding and pooling in addition to the classification process. In[3,7,8] the codebook is constructed using k-means clustering algorithm. Then, each feature vector is quantized to the closest word in the codebook. Regarding the classification process, support vector machine (SVM) is used in,[3,8] whereas weighted nearest neighbor approach is used in.[7] In[9] sparse coding approach and SVM are employed.

A soft and semi-soft coding approach with max-pooling and SVM are used in.[4,5] Two mid-level features, i.e., BossaNova and Fisher vector are used in[6] to improve the accuracy over the classical BoVW approach presented in.[4,5] In[10] the codebook is created using spatial pyramid approach and the classification process is performed by random forest classifier.

## 3.2 Convolutional Neural Networks

Convolutional Neural Networks or Convolutional networks (CNN or ConvNets) have been successfully used in large-scale image and video recognition.[11–14] This success is owing to the availability of a) high-performance GPUs or b) large-scale distributed clusters[15] and c) large public image datasets such as ImageNet.[16] ImageNet Large-Scale Visual Recognition Challenge (ILSVRC) had a great impact on deep visual recognition architectures[17] as it was the main testing platform for large-scale image classification systems.[11,14,18]

CNN was used for exudate detection.[19] Their approach was inspired by[20] where the deep neural networks were used to segment neuronal membranes in electron microscopy. In order to classify each pixel as exudate or non-exudate,[19] used CNN to calculate the probability of the pixel belonging to one of the mentioned classes. The CNN is first trained using several images from a training set and then applied to each pixel in order to segment the image. Hence, the output is a probability image of having a pixel belonging to an exudate or not. Afterward, a fixed threshold is applied to binarize the output. Finally, the optic disc area is masked out.

Pratt et al.[21] solved the five-class problem for the national screening of DR using a CNN. They developed a deep network architecture and data augmentation which can identify the intricate fea-tures involved in the classification task such as micro-aneurysms, exudate, and hemorrhages on the retina and consequently provide a diagnosis automatically without user input. Maninis et al.[22] used a unified CNN for retinal image analysis that provides both retinal vessel and optic disc segmentation. Authors leveraged the latest progress on deep learning to advance the field of automated retinal image interpretation. It requires a single forward pass of the CNN to segment both the vessel network and the optic disc. Gulshan et al. used[23] Inception-v3 CNN architecture[24] with pre-trained ImageNet weights to train and output multiple binary predictions over different grades of diabetic retinopathy. Costa and Campilho[25] presented bag-of-visual-words (BOVW) model on the top of two feature-based CNNs to perform a binary classification (normal and pathological) for diabetic retinopathy diagnosis. Gargeya and Leng[26] developed a gradient boosting classifier over trainable CNN model alongside with prior information (original image meta-data from large-scale datasets) to classify between healthy and abnormal cases in diabetic retinopathy.

## 4 Methodology

In the following subsections, we describe in details the state of art (classical Bag of Visual of Words) proposed approaches (based on deep CNN-features) for retinal classification, see Fig. 2.

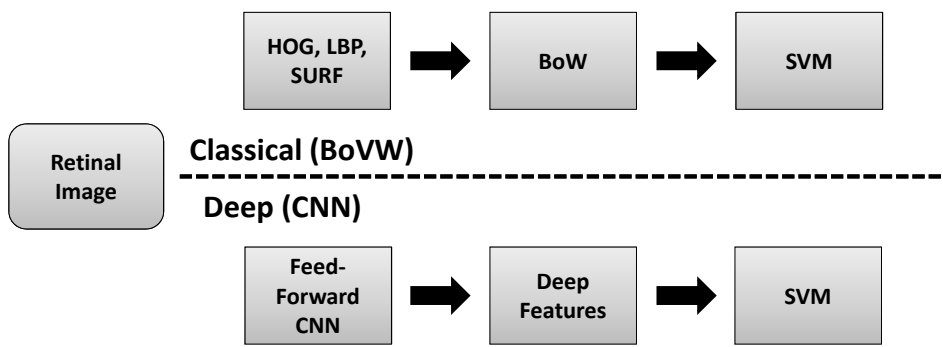

Fig 2: Classical vs deep classification approaches

*4.1 Bag of Visual Words Approach (BoVW)*

Initially, all images are resized to a fixed dimension of 224 × 224 pixels via bilinear interpolation, then the three channels of the RGB color space are separated. An intensity normalization approach is applied to the green channel as discussed in[8] to compensate for the inter-and-intra-patient variability. The normalization step is not applied to other channels, i.e. red and blue, because in some datasets such as STARE the information would be probably lost. Secondly, three local feature descriptors, i.e., Speeded-Up Robust Features (SURF), Histogram of Oriented Gradients (HOG), and Local Binary Patterns (LBP) are extracted from each channel respectively as proposed in.[8] The SURF interest points are detected using the determinant of the Hessian matrix.

The dimension of the SURF descriptors is (64 × total number of interest points) extracted from the three channels as shown in (Fig. 3)(a) and (Fig. 3)(b). The HOG and LBP descriptors are extracted from (28 × 28) pixel patches (non-overlapping) uniformly distributed over the image. The dimension of HOG and LBP descriptors are (64 × 93) and (64 × 174) for the three channels. Each image is represented by a feature matrix encompassing SURF, HOG, and LBP descriptors.

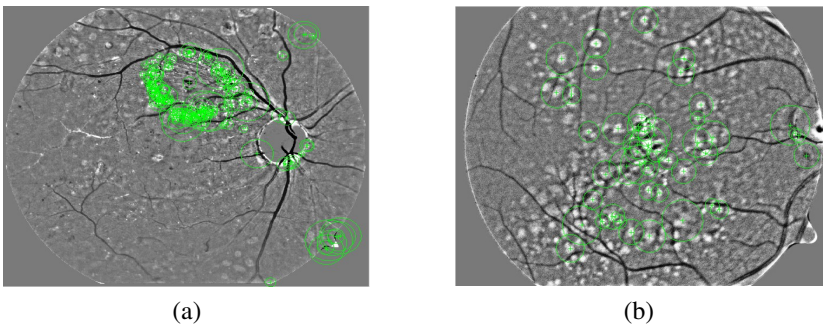

(a)              (b)

Fig 3: SURF interest points of two images containing exudates and drusen respectively.

The codebook is construed using k-means clustering algorithm, assume the set of features extracted from the training sets can be expressed as $\{x_1, x_2, \ldots, x_M\}$ where $x_m \in \mathbb{R}^D$, the idea is to partition this feature set into K clusters $\{d_1, d_2, \ldots, d_K\}, d_k \in \mathbb{R}^D$. The cluster centers represent visual words within the codebook. Each single feature is quantized to its closed word in the codebook using nearest neighbor approach. Finally, the entire image is replaced by a histogram counting the frequencies of words in the codebook. The final histogram feature is $l2$ normalized then fed into a support vector machine with a linear kernel for classification.

*4.2 Deep CNN-Features Approach*

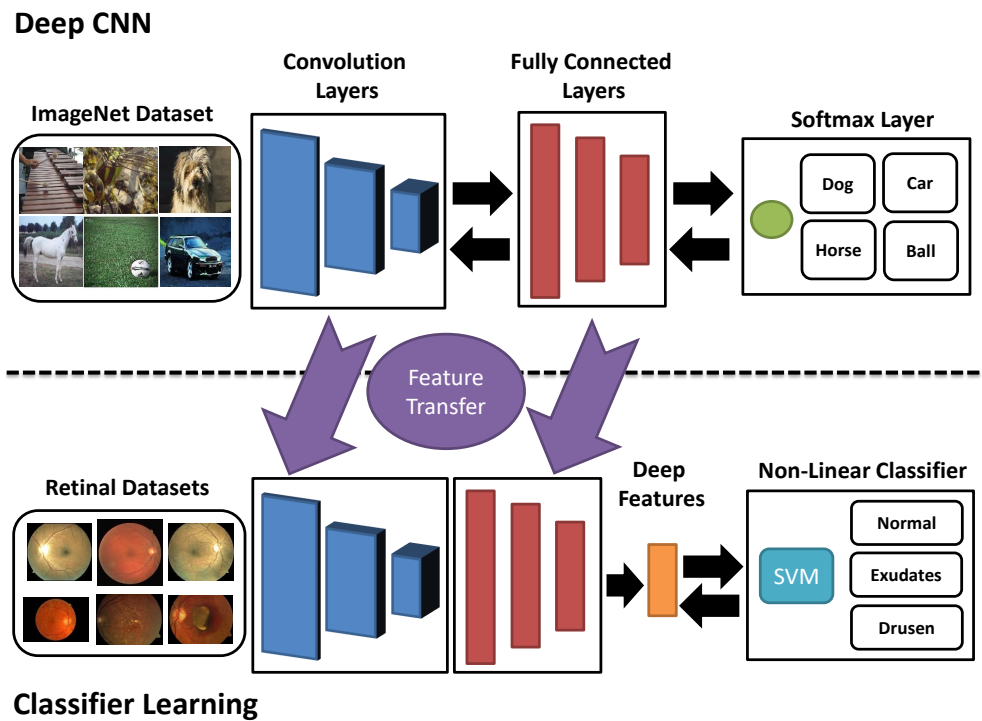

Fig 4: The proposed framework.

As shown in Fig. 4, our method is inspired by feature extraction using very deep trained CNN trained on imageNet,[14] and feature classification using non-linear SVM. The color retinal image is resized to $224 \times 224$ due to the constraints of CNN architecture. The image information is fed forward once through the pre-trained CNN to generate generic features from one of the last fully-connected layers; just before converging the network architecture into domain specific response using softmax layer. The feature vector is further normalized to unit length and classified using a support vector machine with radial basis kernel.

(Fig. 5, Fig. 6, Fig. 7) show detailed feature representations after primary convolution and normalization processes. VGG-VD network[14] don't use any sub-sampling operation during the first layer so that these representations can be visually analyzed to identify the existence of a disease using naked eye (same dimension size as the input image).

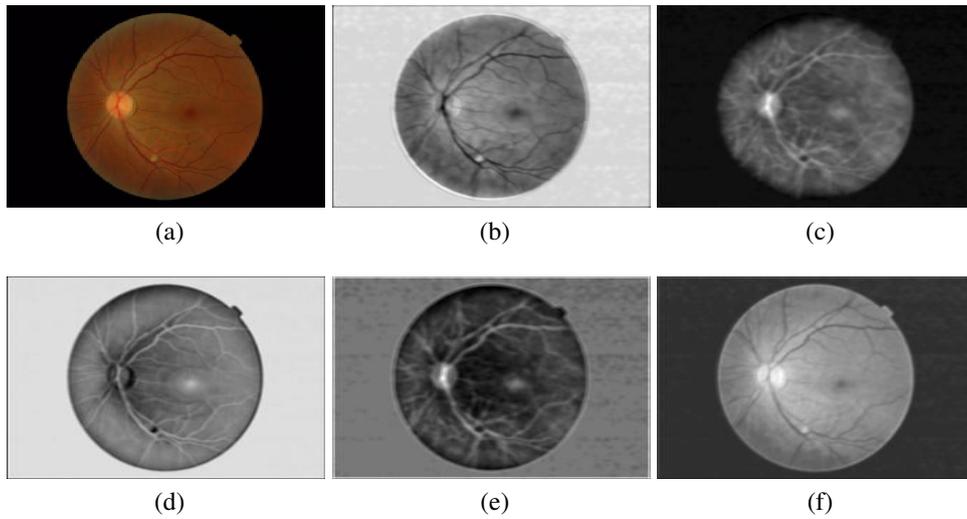

Fig 5: Some feature maps in first layer representing a normal example, using VGG-VD.[14]

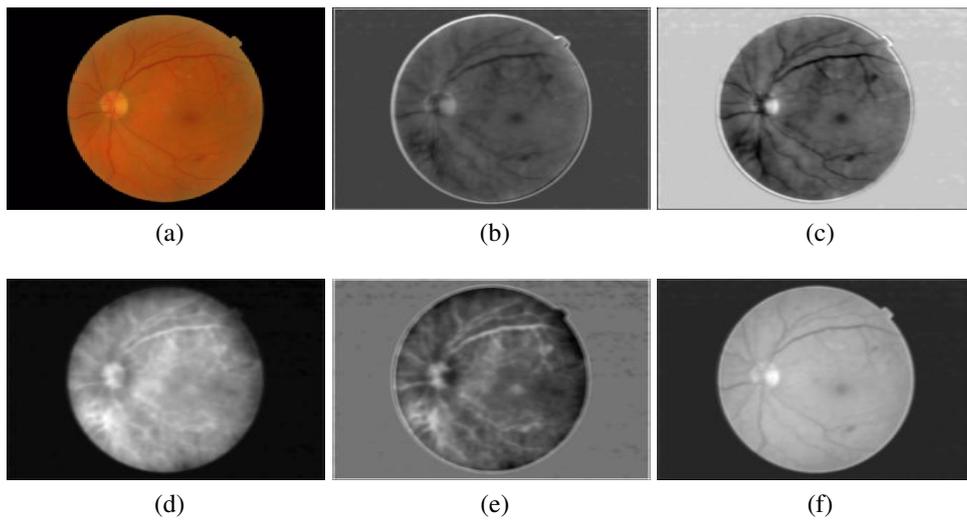

Fig 6: Some feature maps in first layer representing an exudates example, using VGG-VD.[14]

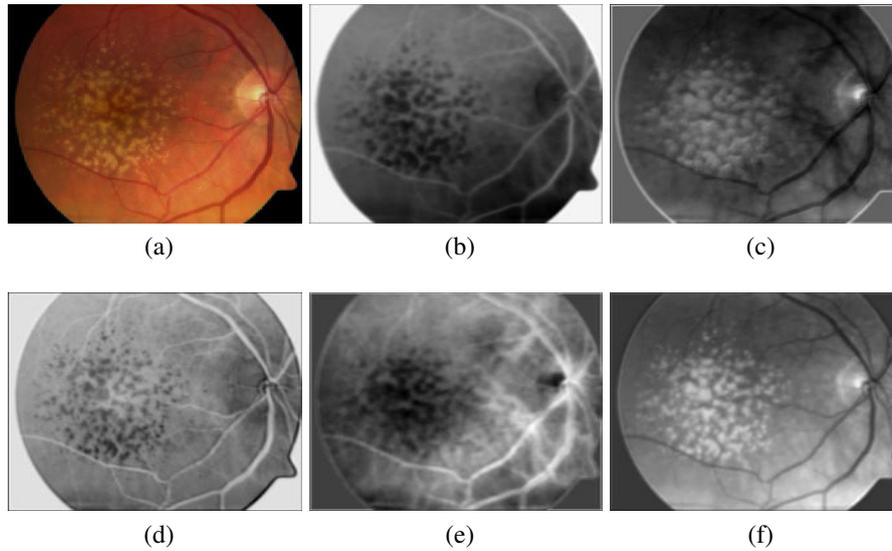

Fig 7: Some feature maps in first layer representing a drusen example, using VGG-VD.[14]

## 5 Results and Discussion

*5.1 Dataset Description*

In the current study, a total of 1133 color retinal fundus images is used with 697 normal images, 352 exudate images, and 84 drusen images as shown in Table 1. Theses images are collected from six publicly available datasets, i.e., STARE[1], HRF[2], DRiDB[3], e_ophtha_EX[4], HEI-MED[5], MESSIDOR[6] in addition to one private dataset[7]. Since the images are collected from different clinical sites, they represent variability in terms of image illumination and contrast. Thereafter, we can assess the robustness of the proposed approach to images acquired from different sources.

---

[1]See: http://www.ces.clemson.edu/~ahoover/stare/.
[2]See:[27] http://www5.cs.fau.de/research/data/fundus-images/.
[3]See:[28] http://www.fer.unizg.hr/ipg/resources/image_database.
[4]See:[29] http://www.adcis.net/en/Download-Third-Party/E-Ophtha.html.
[5]Kindly provided by Luca Giancardo from MIT.[30]
[6]See:[31] http://www.adcis.net/en/DownloadThirdParty/Messidor.html.
[7]Private dataset kindly provided by T. Karnowski from ORNL.

Table 1: Dataset distribution acquired from several datasets.

| Dataset | ORNL | STARE | HRF | DRiDB | e_ophtha_EX | HEI-MED | MESSIDOR | All |
|---|---|---|---|---|---|---|---|---|
| Normal | 36 | - | 15 | 10 | 35 | 61 | 540 | 697 |
| Exudates | 20 | - | - | 28 | 47 | 28 | 229 | 352 |
| Drusen | 61 | 23 | - | - | - | - | - | 84 |

*5.2 Experimental Settings*

The HOG and LBP descriptors are implemented thanks to VLFeat open source library.[32] We chose the recent deep network models (Oxford VGG `imagenet-vgg-m-2048`,[33] Oxford VGG-VD `imagenet-vgg-verydeep-19`,[14] GoogLeNet `imagenet-googlenet-dag`,[34] and MicroSoft ResNet `imagenet- resnet-50-dag`[35]) from MatConvNet library[36] to extract CNN feature vectors, which are pre-trained on ImageNet dataset.[16] The LIBSVM machine learning open source library[37] is used to solve the classification problem, in which linear and radial basis kernel functions are selected for BoVW and deep CNN-features respectively. We applied this classification algorithm with a penalty term. i.e., constant $C = 8$ and a 10-fold cross validation. This value of the constant $C$ minimizes the percentage of classification errors as discussed in.[8] The MATLAB source code of this research paper using deep CNN method based on GoogLeNet model was made publicly available at https://github.com/mawady/DeepRetinalClassification. The objective of the 10-fold cross validation is to split the data into 10 subsets, and the fitting process is performed with 9 of the subsets. However, the classification is performed via the remaining subset. The entire process is repeated 10 times to make sure that each image is included for both training and testing. Finally, the mean classification performance is recorded. We compute the mean accuracy, sensitivity, and specificity for each individual class (normal, exudates, and drusen) to appraise the behavior of the proposed approach.

*5.3 Performance Analysis*

Regarding BoVW approach, the statistical measures are calculated at different visual words and the mean values are then computed for each visual word. As presented in Table 2, the codebook is constructed at five different values such as $W = [100, 200, 300, 400, 500]$. The highest Acc and Spec for the drusen class are achieved at W = 400 such as Acc = 99.65% and Spec = 100% respectively. The highest system results are conducted at W = 500, where Acc measurements for normal and exudates are (79.79% , 80.05%), Sens measurements for normal, drusen, and exudates are (80.97%, 71.79%, 98.75%), and Spec measurements for normal and exudates are (77.96%, 83.13%) respectively. In the deep CNN-features approach, different deep network models were computed for statistical comparison, and the mean measurements are produced individually for each model. As shown in Table 3, the Sens measurements of exudates and the Spec measurements of normal increase as more deeper the network employed (number of inner layers for ResNet,[35] GoogLeNet,[34] VGG-VD[14] and VGG[33] are 515, 153, 42 and 21 respectively). The best feature results are extracted from GoogLeNet model, where the Acc, Sens and Spec measurements for normal and exudates are highest. Averaged across the three classes, we are able to achieve an accuracy of 92.00%, a sensitivity of 87.79%, and a specificity of 90.67%.

In general, the two approaches have good performance with the drusen class. However, the BoVW approach have a lower performance compared to the deep features approach regarding the normal and exudate classes. This problem originates because the intensity of the optic disc is very close to exudate lesions.[38] Moreover, some of the exudate lesions in the MESSIDOR dataset are very difficult to detect. i.e, the lesions are very small or less distributed comparing to other exudate lesions in the dataset.

Table 2: The mean accuracy (Acc), sensitivity (Sens), and specificity (Spec) of the 10-fold cross validation results for the BoVW method. Norm: normal, Ex: exudates, Dru: drusen, W: visual words, Max: maximum.

|  | Acc (%) | | | Sens (%) | | | Spec (%) | | |
| --- | --- | --- | --- | --- | --- | --- | --- | --- | --- |
| W | Norm | Ex | Dru | Norm | Ex | Dru | Norm | Ex | Dru |
| 100 | 75.02 | 75.37 | 98.77 | 74.20 | 69.42 | 88.98 | 77.72 | 76.49 | 99.71 |
| 200 | 77.14 | 77.32 | 99.12 | 77.39 | 68.98 | 95.46 | 76.75 | 79.69 | 99.44 |
| 300 | 76.96 | 77.05 | 99.21 | 78.58 | 66.24 | 92.71 | 73.81 | 80.55 | 99.81 |
| 400 | 78.63 | 78.81 | **99.65** | 79.68 | 70.17 | 95.78 | 76.51 | 81.65 | **100.00** |
| **500** | **79.79** | **80.05** | 99.56 | **80.97** | **71.79** | **98.75** | **77.96** | **83.13** | 99.62 |
| Max | 79.79 | 80.05 | 99.65 | 80.97 | 71.79 | 98.75 | 77.96 | 83.13 | 100.00 |

Table 3: The mean accuracy (Acc), sensitivity (Sens), and specificity (Spec) of the 10-fold cross validation results for the CNN models. Norm: normal, Ex: exudates, Dru: drusen.

|  | Acc (%) | | | Sens (%) | | | Spec (%) | | |
| --- | --- | --- | --- | --- | --- | --- | --- | --- | --- |
| Models | Norm | Ex | Dru | Norm | Ex | Dru | Norm | Ex | Dru |
| VGG | 86.59 | 86.59 | **99.82** | 95.90 | 64.82 | 98.22 | 71.58 | 96.26 | **100.0** |
| VGG-VD | 85.41 | 85.14 | 99.74 | 94.08 | 67.27 | **99.00** | 74.25 | 94.59 | 99.80 |
| **GoogLeNet** | **88.17** | **88.17** | 99.65 | **96.02** | textbf70.36 | 97.00 | **75.79** | **96.33** | 99.90 |
| ResNet | 86.94 | 86.94 | **99.82** | 94.27 | 69.47 | 98.75 | 75.35 | 94.89 | 99.91 |

Therefore, the system miss-classifies exudate images as normal and vice versa. Fig. 8a shows an example of normal image. While, Fig. 8b and Fig. 8c show two examples for exudate images. In the first, the exudate lesions are less distributed as indicated inside red rectangle. In the latter, the lesions are visible and more distributed.

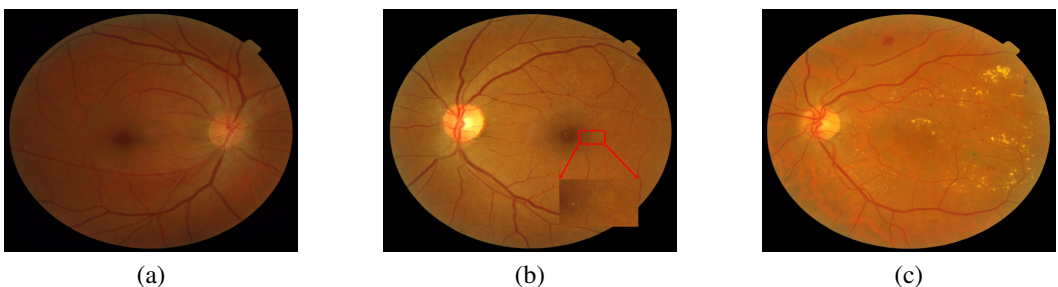

(a)     (b)     (c)

Fig 8: Image examples from MESSIDOR dataset.

In[9] authors achieved accurate results for the three classes. However, it should be noted that authors don't use all normal and exudate images from the MESSIDOR dataset. i.e., (351 normal and 218 exudates). In our case, we already used all normal and exudate images (540 normal and 229 exudates) according to the rules mentioned in [8] In Table 4, we compare the proposed methods (classical: BoVW and deep: VGG, VGG-VD, GoogLeNet, ResNet). Deep approaches achieve better results than classical one. GoogLeNet has the best average accuracy among others, while ResNet has a more stable performance with lowest accuracy variance.

Table 4: Overall performance of the proposed approaches with SVM classification

|  | BoVW | VGG | VGG-VD | **GoogLeNet** | ResNet |
|---|---|---|---|---|---|
| Accuracy | $77.76 \pm 1.97$ | $91.83 \pm 2.93$ | $90.76 \pm 1.93$ | $\mathbf{92.00 \pm 1.53}$ | $91.23 \pm 1.07$ |

## 6 Conclusion

This work proposed to automatically extract deep features from 4 different CNN models for 3-class retinal classification. The method was trained and tested on 6 highly different retinal datasets with cross validation. Experimental results showed superior performance over conventional BOW approach by more than 10% increase in mean accuracy. In future work, the proposed approaches can be improved by extracting feature vectors from the outputs of different convolutional layers which dimensionality will be further reduced to a more compact form by using standard analysis approaches (e.g PCA).

**Conflict of interest statement**

The authors declare no competing financial interests.

---

[8] http://www.adcis.net/en/DownloadThirdParty/Messidor.html.

**Ibrahim Sadek** is a PhD student at Télécom ParisTech, Institut Mines Télécom, France. In 2010, he received a B.Sc. in biomedical engineering from biomedical engineering dept., Helwan University, Egypt. In 2014, he received an Erasmus Mundus joint master degree in computer vision and robotics (VIBOT). During his master, he has attended three different universities. i.e., University of Burgundy (Dijon, France), University of Girona (Girona, Spain), and heriot watt university (Edinburgh, Scotland). He has done his master thesis at Le2i, UMR CNRS 6303, University of Burgundy (Dijon, France). He received the Robert F. Wagner All Conference Best student paper award for my paper "Automatic Discrimination of Color Retinal Images Using the Bag of Words Approach" published at SPIE medical imaging conference on computer aided diagnosis, February 2015. Orlando - Florida USA.

Biographies and photographs of the other authors are not available.

# List of Figures

## List of Tables